%% file: main.tex
\documentclass[letterpaper, 10 pt, conference]{ieeeconf}  

\IEEEoverridecommandlockouts                              

\usepackage{graphicx} 
\usepackage{algorithm}
\usepackage{multirow}
\usepackage{amssymb}
\usepackage{amsmath} 
\usepackage{cite}

\usepackage{soul}
\usepackage{xcolor}
\soulregister{\cite}7
\soulregister{\citep}7
\soulregister{\citet}7
\soulregister{\ref}7
\soulregister{\pageref}7
\soulregister{\uppercase\expandafter}7
\sethlcolor{yellow}
\title{\LARGE \bf
Building Explicit World Model for Zero-Shot Open-World Object Manipulation
}

\author{Xiaotong Li, Gang Chen, and Javier Alonso-Mora
\thanks{All authors are with the Department of Cognitive Robotics (CoR), Delft
University of Technology, 2628CD Delft, The Netherlands.
{\tt\small X.Li-89@student.tudelft.nl; 
 \{g.chen-5;
j.alonsomora\}@tudelft.nl}}%
}

\begin{document}

\maketitle
\thispagestyle{empty}
\pagestyle{empty}

\input{Content/abstract}
\input{Content/introduction}
\input{Content/related_works}
\input{Content/methods}
\input{Content/results}
\input{Content/discussion}

\bibliographystyle{IEEEtran}

\bibliography{bib/final}

\end{document}

%% file: Content/abstract.tex
\begin{abstract}
Open-world object manipulation remains a fundamental challenge in robotics. 
While Vision–Language–Action (VLA) models have demonstrated promising results, they rely heavily on large-scale robot action demonstrations, which are costly to collect and can hinder out-of-distribution generalization.
In this paper, we propose an explicit-world-model-based framework for open-world manipulation that achieves zero-shot generalization by constructing a physically grounded digital twin of the environment. The framework integrates open-set perception, digital-twin reconstruction, sampling and evaluation of interaction strategies. By constructing a digital twin of the environment, our approach efficiently explores and evaluates manipulation strategies in physic-enabled simulator and reliably deploys the chosen strategy to the real world.
Experimentally, the proposed framework is able to perform multiple open-set manipulation tasks without any task-specific action demonstrations, proving strong zero-shot generalization on both the task and object levels. 
Project Page: https://bojack-bj.github.io/projects/thesis/
\end{abstract}

%% file: Content/introduction.tex
 \section{Introduction}
Open-world manipulation has recently become a popular research frontier in robotics, aiming to enable robots to perform diverse tasks commanded by humans in unstructured environments. In such settings, robots must have the ability to infer task goals from natural language, perceive and interact physically with previously unseen objects. This open-ended nature introduces fundamental challenges in semantic and physical understanding.



Recent works have explored Vision–Language–Action (VLA) models~\cite{kim2024openvla, ghosh2024octo, brohan2023rt2visionlanguageactionmodelstransfer, black2024pi0visionlanguageactionflowmodel}, which leverage large-scale vision–language backbones plus an action head to predict robot actions for diverse tasks.
Despite strong performance on manipulation benchmarks, VLAs often struggle to generalize beyond the training distribution and rely on large amounts of costly robot demonstration data for supervision.
Another line of research investigates world models, which predict the consequences of actions rather than directly imitating demonstrations, either through implicit dynamics modeling~\cite{liao2025genie} or explicit world construction~\cite{jiang2025dexsim2real2buildingexplicitworld}. By reasoning over action outcomes, world-model-based approaches offer improved generalization to unseen objects and tasks.

Most existing world models are image- or video-based~\cite{hafner2022masteringataridiscreteworld,yang2023learning,bruce2024geniegenerativeinteractiveenvironments}. While these 2D models can generate visually compelling predictions, they typically lack explicit 3D structure and physical constraints, which limits their ability to faithfully capture real-world dynamics and physical laws. In contrast, explicit world models represent the environment with geometrically and physically meaningful digital-twin assets~\cite{barcellona2024dream}. However, prior explicit-world-model approaches either target specialized problem settings, such as articulated-object manipulation toward desired joint configurations~\cite{jiang2025dexsim2real2buildingexplicitworld}, or still rely on task-specific demonstrations to train policies~\cite{barcellona2024dream}. As a result, a method that can generalize zero-shot to both novel objects and novel tasks remains needed.




In this paper, we propose an explicit world-model-based framework for object manipulation in open-world environments. In such settings, target objects are not predefined, and the robot must operate on previously unseen, open-set objects. A central challenge is therefore to accurately reconstruct a digital twin from onboard 2D observations, which are inherently partial and can be further degraded by occlusions, e.g., from the gripper during grasping.
To address this challenge, our framework combines modern generative and visual foundation models with classical point-cloud registration techniques to reconstruct digital twins that are both geometrically accurate and semantically consistent. Given a high-level task specification, we then sample diverse interaction strategies within the reconstructed world model. A physics engine is used to simulate the outcomes of these candidate strategies, and a large vision–language model (VLM) evaluates their consistency with the task prompts to select the most promising action sequence.
The entire pipeline operates without any task-specific training and is independent of robot embodiment, enabling zero-shot generalization to novel objects and tasks.

The contributions of this work are listed as follows:
\begin{itemize}
    \item An explicit world model based manipulation framework that achieves zero-shot generalization to novel rigid objects and task specifications, without task-specific training or demonstrations.
    \item A dynamic digital twin construction pipeline that reconstructs object meshes from onboard observation, and aligns scales and poses to real objects for simulation-consistent transfer.
    \item A VLM-based evaluation module that evaluates the success probabilities of simulated outcomes of sampled action candidates.
\end{itemize}
We validate our system by evaluating digital-twin construction accuracy and conducting real-robot experiments on multiple open-set, semantically subtle manipulation tasks with previously unseen objects, reporting both qualitative and quantitative results.


%% file: Content/related_works.tex
\section{related works}

\subsection{Open-world manipulation}
Manipulation has long been a popular topic in robotics due to its substantial practical value. Traditional reinforcement learning \cite{levine2016end} and imitation learning \cite{chi2025diffusion,zhao2023learning} methods have achieved good performance in this domain, but they typically require training a separate policy for each task, and thus exhibit limited generalization. In contrast, open-world manipulation demands a system that can generate policies for a broad, open-ended set of tasks. 

In the early stage, researchers explored leveraging pretrained vision–language foundation models to achieve strong generalization. 
A common paradigm is to use a foundation model to infer task-specific affordances \cite{tang2024uad} or a coarse trajectory/keypoints \cite{yuan2025robopoint} for open-ended instructions. 
In \cite{huang2025rekep,liu2024moka}, the authors both choose to use VLM to predict 2D keypoints, and lift to 3D trajectories. While this approach can generalize well, the task success depends heavily on the accuracy of the model predictions. It typically lacks closed-loop verification and real-time reaction to ensure that the desired outcome is achieved. Thus, in more complete scenarios, this purely feedforward fashion can break down.

Motivated by these limitations, subsequent work has moved toward Vision–Language–Action (VLA) models.
VLA models typically adopt a pretrained VLM as the backbone to encode observations and natural-language task instructions, with additionally an action head to produce executable control commands or policies.
This design is architecturally concise and can directly inherit strong semantic priors from large pretrained VLMs.
However, action data is a fundamentally different modality from language and vision, and a backbone pretrained primarily on discrete tokens of 2D vision and language does not naturally handle this continuous data representation.
As a result, VLA performance and robustness remain strongly bottlenecked by the scale and quality of action supervision.

Recently, the emergence of models such as Pi\cite{black2024pi0visionlanguageactionflowmodel} and the release of large-scale Open X-Embodiment \cite{o2024open} datasets have signaled a community-wide effort to scale up VLA training data and push toward an actionable scaling law.
Nevertheless, collecting diverse, high-quality robot action trajectories remains expensive and difficult to standardize across embodiments and sensing configurations. Fully realizing the benefits of scaling-up will likely take time.

\subsection{World Models}
World models were originally introduced in reinforcement learning to improve sample efficiency by enabling agents to \emph{imagine} the consequences of interactions in a learned latent space ~\cite{hafner2022masteringataridiscreteworld,hafner2020dreamcontrollearningbehaviors,wu2022daydreamerworldmodelsphysical}.
By encapsulating the environment dynamics, a world model allows an agent to roll out hypothetical futures, support planning, and make informed decisions without executing every action in the real world.

With the recent surge of video diffusion models, much of the progress has focused on image- or video-based world models~\cite{yang2023learning, bruce2024geniegenerativeinteractiveenvironments}.
Despite producing visually compelling predictions, such 2D world models often struggle to provide explicit 3D structure and physical consistency, due to their 2D training data and underlying 2D representations.
This limitation makes it difficult to faithfully reason about geometry, contact, and dynamics required for manipulation.

In parallel, another line of research constructs \emph{explicit} world models by leveraging 3D reconstruction or generation techniques, typically focusing on object-centric reconstructions of the entities involved in the task.
In \cite{barcellona2024dream}, authors propose an explicit-world-model framework that reconstructs a sim-ready digital twin from visual observations and leverages physics-based simulation to train imitation learning policy. Jiang et al.~\cite{jiang2025dexsim2real2buildingexplicitworld} construct explicit world models for articulated objects and plan trajectories via sampling-based model predictive control within a physics simulator, without requiring demonstrations or reinforcement learning.

Although these approaches also leverage explicit world models, our focus is different. Compared with \cite{jiang2025dexsim2real2buildingexplicitworld}, which focuses on articulated-object manipulation toward target joint configurations, our framework supports a wider range of open-set manipulation tasks beyond articulated joint-angle goals. In contrast to \cite{barcellona2024dream}, which still relies on task demonstrations to train imitation-learning policies, our method enables zero-shot generalization to previously unseen tasks, without requiring any task-specific demonstrations.

%% file: Content/methods.tex
\section{Methods}

\subsection{Problem Statement}

Formally, given the visual observations $I$ of the scene (e.g., RGB-D images), and a natural language task instruction $C$, the objective is to build a world model $W$, and find an $a$ that successfully achieves the goal described by $C$ in the real world.
That is, we first build a world model $W$
\begin{equation}
    W:p_W(\tau \mid I,a),
\end{equation}

where $\tau=(s,o)$ is the future states and observations generated by world model $W$, given the current observation $I$ and sampled action $a$. Then,
\begin{equation}\label{eqa1}
    \begin{aligned}
        a^* = \arg\max_{a \in \mathcal{A(I)}} E_{\tau\sim p_W(\cdot \mid I, a)}[R(C,\tau)], \\
        \mathcal{A} = \{a_i\} \sim \pi(\cdot \mid I,C), \; i=1,\ldots,N
    \end{aligned}
\end{equation}
where $\mathcal{A}$ denotes the space of sampled candidate actions from a sample policy $\pi(\cdot \mid I,C)$, parameterized by end-effector poses, $a^*$ is the selected best action and $R(C,\tau)$ is a score function that measures the score of $\tau$ given the natural language task instruction $C$. 

In this paper, we aim to build an explicit world model that can generate plausible future states and observations $\tau$. We also designed a prior-based sample policy $\pi$ and a VLM-based result checker to measure the success score of $\tau$.
We focus on rigid objects. Handling deformable and articulated objects would require incorporating elasticity modeling, joint constraints, and contact deformation into the world model. We leave these extensions to future research.

\begin{figure*}[t]
    \centering
    \includegraphics[width=0.95\hsize]{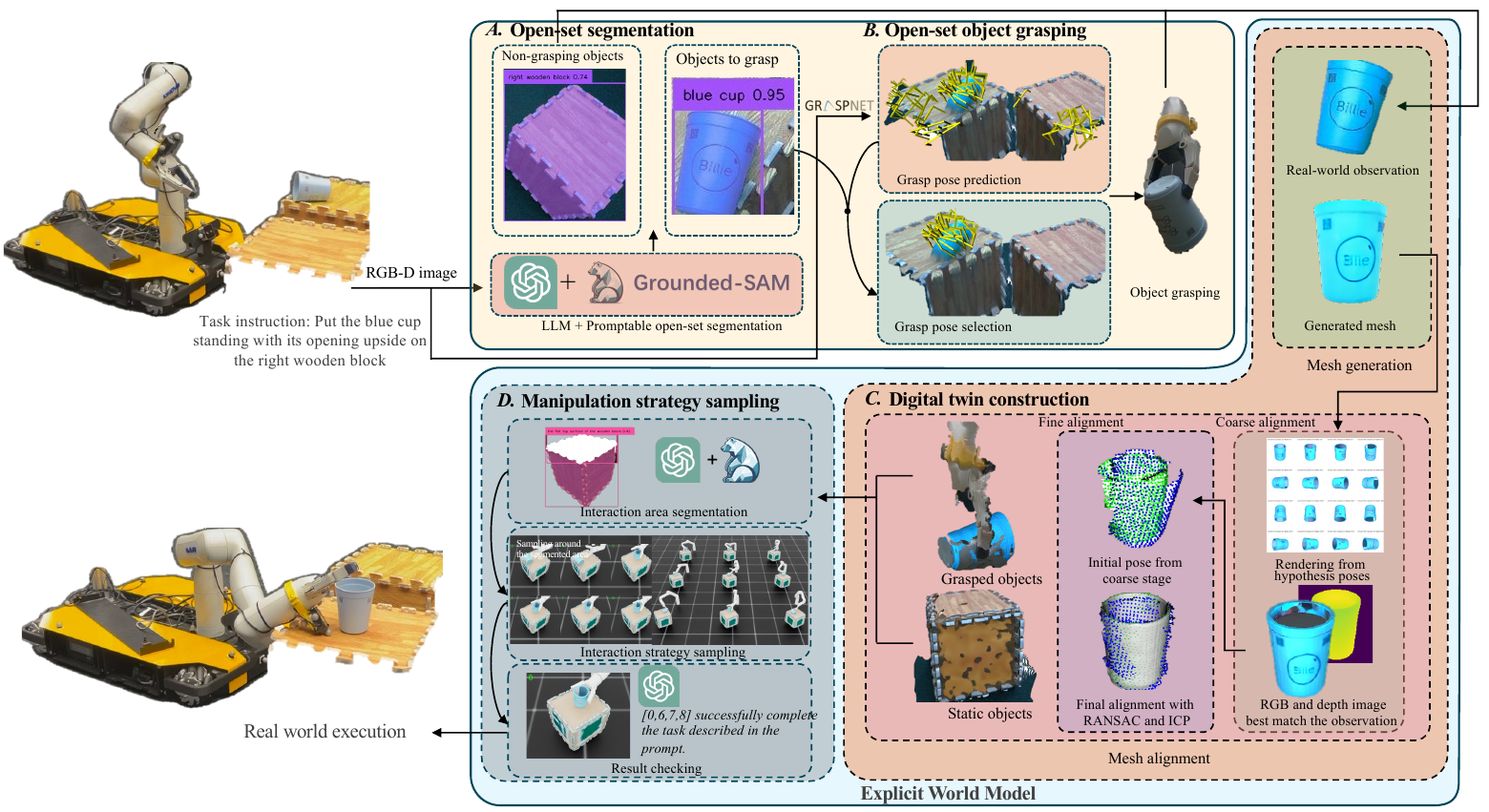}
    \caption{Overview of our manipulation task framework with explicit world model construction. The system takes an RGB-D observation $I$ and a natural language task instruction $C$ as input and proceeds through four main stages: (A) \textit{Open-set segmentation}, (B) \textit{Open-set object grasping}, (C) \textit{Digital twin reconstruction}, (D) \textit{Manipulation strategy sampling}. The outcome is the most promising manipulation strategy, which is then executed on the real robot.}
    \label{fig:pipeline}
\end{figure*}

\subsection{System Overview}
An overview of the full framework is illustrated in Figure~\ref{fig:pipeline}. Given an RGB-D image of the scene and a text instruction (e.g., “put the blue cup standing with its opening upside on the right wooden block”), the system outputs a complete manipulation plan and executes it on a real-world manipulator. To achieve our goal, the framework consists of four main components: \textit{Open-set Segmentation and Grasping} (Section~\ref{sec:grasp}), \textit{Digital Twin Construction} (Section~\ref{sec:digital}), and \textit{Manipulation Strategy Sampling} (Section~\ref{sec:sampling}).

\subsection{Open-set Segmentation and Grasping}\label{sec:grasp}
In this module, we first employ GPT-4o~\cite{openai2024gpt4ocard} and Grounded-SAM~\cite{ren2024groundedsamassemblingopenworld} to segment any objects related to the given task prompt. Specifically, we query the GPT-4o with the RGB image and a prompt question (e.g., “What objects in the picture are involved in the task ‘put the blue cup standing with its opening upside on the right wooden block’? Return their names in two categories as directly manipulated objects and other interactive objects.”). The VLM here allows our method to generalize to arbitrary objects beyond a closed set of predefined classes. Once the object are identified, we use Grounded-SAM to segment the target objects from the input image. 
This gives us accurate 2D segmentation masks $\Tilde{M}_\text{obj}$ for both directly manipulated objects (e.g., cup) and other interactive objects involved in interaction (e.g., right wooden block).


For object grasping, we adopt AnyGrasp~\cite{fang2023anygrasprobustefficientgrasp} as our grasp pose prediction module, which is trained on a large number of real-world grasping data and has demonstrated strong robustness in open-set, general-purpose grasping tasks. The grasp pose predictor, denoted as $f_g(\cdot)$, takes an input RGB-D image $I$ and outputs a set of grasp pose candidates $ \Tilde{g} =  f_g(I)$.
The predicted grasp candidates \( \Tilde{g} \) are densely distributed among all graspable objects in the scene. To reduce computational overhead, we first retain only the top 1000 predictions ranked by confidence. For instance-level grasping, we reuse the masks \( \Tilde{M}_{\text{obj}} \) of the target objects obtained from the open-set segmentation module, retaining only those associated with the object of interest. Given the $\Tilde{M}_\text{obj}$ of the target objects, we perform back-projection on the masked RGB-D image $I_{obj}$ to obtain the corresponding partial 3D point cloud $P_{\text{obj}}$. Each grasp candidate in \( \Tilde{g} \) is then evaluated based on its spatial proximity to the surface of $P_{\text{obj}}$. Specifically, we retain only those grasp poses which lie within a predefined distance threshold from the segmented object’s surface points, 
\begin{equation}
s(g) \;=\;
\begin{cases}
1, & \text{if } d(g)\le \tau \quad (\text{select})\\[4pt]
0, & \text{if } d(g)>\tau \quad (\text{discard})
\end{cases}
\end{equation}
\begin{equation}
d(g) \;\triangleq\; \min_{\,\mathbf{p}\in \mathcal{P}_{\text{obj}}} \bigl\| \mathbf{x}(g)-\mathbf{p} \bigr\|_2
\qquad
\end{equation}

\begin{figure*}[t]
    \centering
    \includegraphics[width=0.9\hsize]{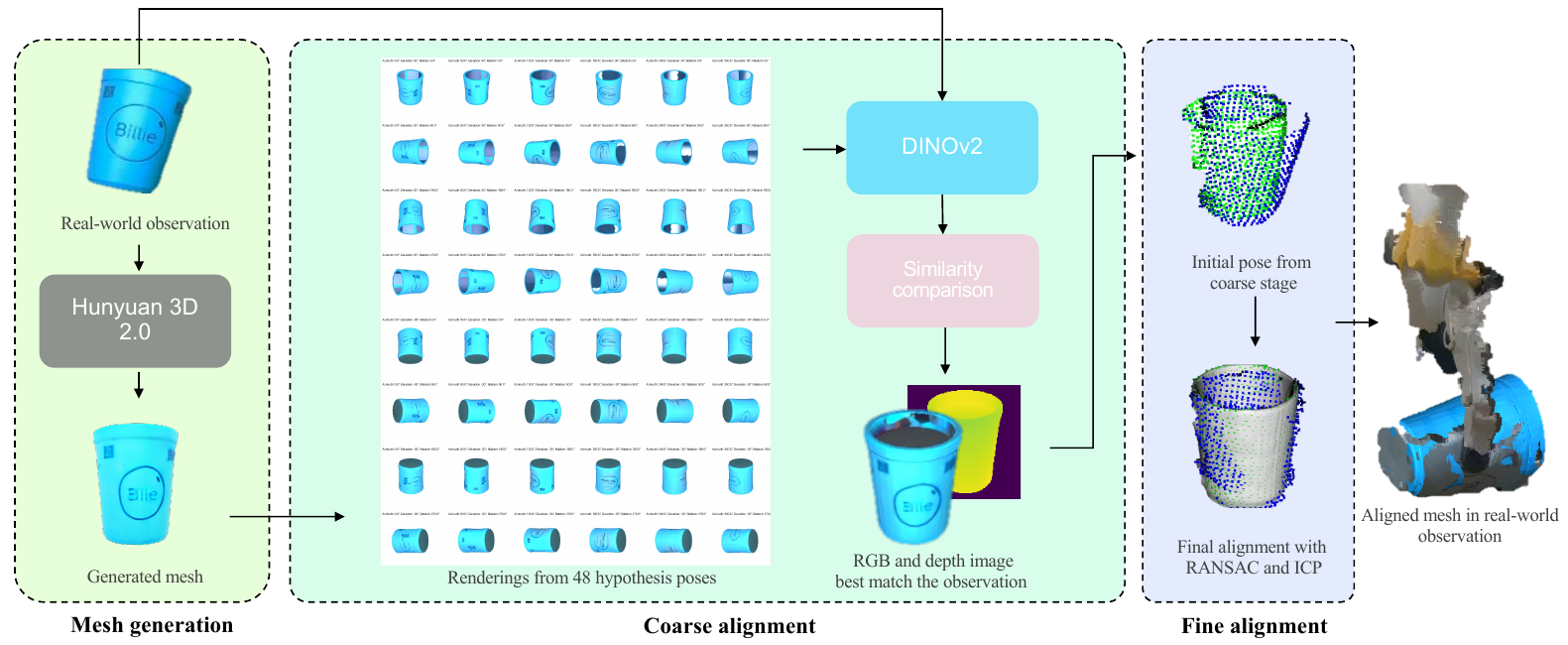}
    \caption{The proposed digital twin construction module, containing the mesh generation and two-stage pose alignment. We first generate a textured mesh from the masked RGB image via Hunyuan3D 2.0~\cite{zhao2025hunyuan3d20scalingdiffusion}. During coarse alignment, we render RGB and depth images from a set of hypopaper poses and compare their similarities with real-world observation in DINO~\cite{oquab2024dinov} feature space, and select the one that best matches the real-world observation. The resulting coarse pose is then refined using RANSAC and ICP on the partial point cloud back-projected from the depth image.}
    \label{fig:align}
\end{figure*}

To improve the robustness of the grasping, we also add a grasp result checker. After the execution of grasping, we inquire the GPT-4o with the observation from the bottom camera if the gripper successfully grasped the target object, and if the answer is no (the grasping fails or someone deliberately takes the object off), the system will keep trying to find the object and grasp it.

\subsection{Digital Twin Construction}\label{sec:digital}
The objective of this module is to construct accurate 3D meshes of the target objects, align them with real-world observations, and predict their material properties, enabling reliable use in simulation-based interaction and planning.


This project adopts Hunyuan 3D 2.0~\cite{zhao2025hunyuan3d20scalingdiffusion} as the 3D generation module. 
However, the mesh generated by the 3D generation module is in unit scale and canonical pose. Therefore, it is necessary to align the generated mesh with the real-world observation by estimating the appropriate scale and 6-DoF pose.
Since only partial object geometry is observed (often limited to a single visible side) and additional occlusions may be introduced by the gripper, classical point cloud based alignment methods tend to perform poorly in this scenario.
To address this, we proposed two-stage pose alignment pipeline, performs mesh-observation matching in a coarse-to-fine manner, as shown in Fig~\ref{fig:align}. 

Let $\Tilde{P}$ denote the set of hypothesized coarse object poses sampled in 6D space (translation and rotation).
For each hypothesized pose $\Tilde{p} \in \Tilde{P}$, the corresponding textured 3D mesh is rendered to obtain both the RGB image $\Tilde{I}\text{rgb}$ and the depth image $\Tilde{I}\text{depth}$.
The following equation describes this rendering process:

\begin{equation}
\Tilde{I}_\text{rgb}, \Tilde{I}_\text{depth} = \text{Render}_{\Tilde{P}}(\text{Mesh}).
\end{equation}

Then, we compare the similarities of these rendered views with the real-world observation $I_{obs}$ to select the most promising hypothesis as an initial coarse pose estimate. Here we use DINOv2~\cite{oquab2024dinov} as the feature extractor and compute the cosine similarities,
\begin{align}
    &\Tilde{F}_{\text{render}} = \text{DINO}(\Tilde{I}_{\text{rgb}}) \\
    &F_{\text{obs}} = \text{DINO}(I_{\text{obs}})\\
    &\Tilde{S} = cosine(\Tilde{F}_{\text{render}}, F_{\text{obs}}),
\end{align}
We then select the hypothesized pose with the highest similarity to the real-world observation as the coarse pose estimate. Using this pose, we render a depth image of the generated mesh and back-project it to obtain a partial point cloud that corresponds to the observed view. This transformation effectively converts the challenging partial-to-complete alignment problem into a partial-to-partial alignment problem, thereby enabling the use of conventional point cloud registration methods, such as RANSAC and ICP, for further fine alignment.



The coarse alignment produces two roughly aligned point clouds that still differ in scale. To estimate the scale factor, we compare the dimensions of the 3D bounding boxes derived from the partial point clouds of both the rendered mesh and the real observation, and adjust the mesh dimensions to match the real-world object. Following this, we refine the transformation using a combination of RANSAC and ICP, yielding an accurate alignment between the generated mesh and the real object. We examined our mesh alignment pipeline and the results are in Sec.~\ref{sec:alignment_exp}.

Moreover, Inspired by Xu et al.~\cite{xu2024gaussianproperty}, who leveraged an LLM together with a curated material library to infer the material and physical properties of an object from its masked image, we utilize the reasoning capability of GPT-4o to predict the material of the object. The predicted material can then be used to assign corresponding physical properties in the simulation environment. 

\subsection{Manipulation Strategy Sampling}\label{sec:sampling}

\begin{figure*}[h]
    \centering
    \includegraphics[width=0.9\hsize]{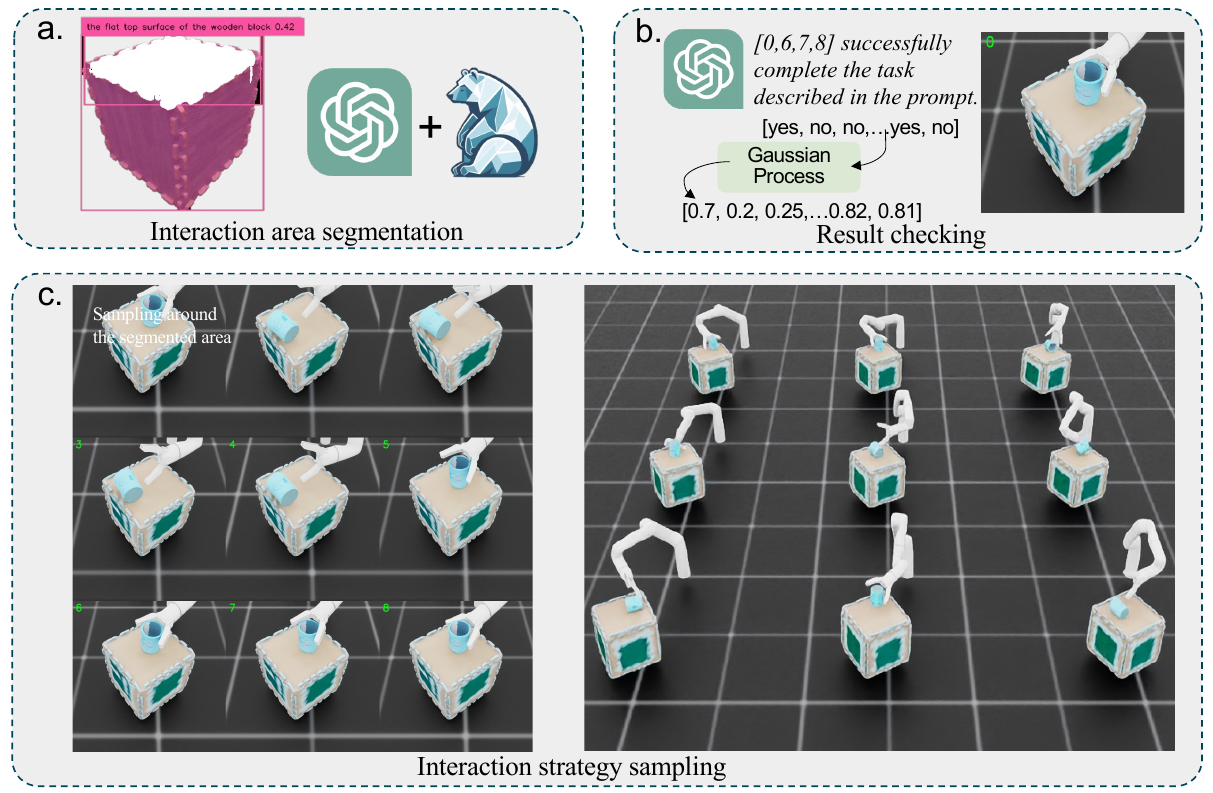}
    \caption{Our manipulation strategy sampling module. (a) \textit{Interaction area segmentation}: We use GPT-4o and Grounded-SAM to segment the interaction area to constrain the sample space of translation. (b) \textit{Interaction strategy sampling}: Different rotations of the dynamic object are sampled. The outcomes are simulated in the Isaac Sim. (c) \textit{Result checking}: The results are rendered in the simulator. We query the GPT-4o again to check which samples fulfill the task requirements, and feed the results to a Gaussian Process classifier to get the success probabilities.}
    \label{fig:sampling}
\end{figure*}

Once the digital twin of the environment is constructed, we integrate it into a physics-enabled simulation environment to perform manipulation strategy sampling. In this work, we adopt NVIDIA Isaac Sim as the simulation platform. We then sample 6-DoF poses for the grasped object, representing candidate manipulation goals. 
The simulated outcome of each sample is then evaluated to determine whether the manipulation strategy successfully completes the task. This process allows us to test diverse candidate strategies without executing them on the real robot.

To improve sampling efficiency and reduce the size of the search space, the translation component of the dynamic object’s pose is constrained to be near the most probable interaction region, which is predicted by GPT-4o plus Grounded-SAM, as illustrated in Fig.~\ref{fig:sampling}a. 
The segmented region is back-projected into 3D space to obtain a point cloud, from which we compute the centroid as reference for the initial estimate of the translation component. 

After constraining the translation around the interaction area, we sample different rotation angles, which are reachable by the robot arm, to generate diverse 6-DoF pose hypotheses. For each sample, we spawn the non-directly manipulated object at its real-world pose and the grasped object at the sampled pose, and the Kinova arm with proper joint values computed by an inverse kinematics solver. 

For each simulated strategy sample, we render an RGB image as the observation for each candidate from a fixed, front-facing viewpoint with a $-60^\circ$ tilt, shown in Fig~\ref{fig:sampling} b on the left. We then query GPT-4o with the rendered image and the natural language task instruction to determine whether the observed outcome satisfies the instruction. This yields weak binary labels $y_i\in\{0,1\}$ for each end-effector pose ($\mathbf{t}_i,\mathbf{q}_i$), where $\mathbf{t}_i\in\mathbb{R}^3$ is translation and $\mathbf{q}_i=[w\;x\;y\;z]^\top$ is a unit quaternion. The strategies deemed successful by the GPT-4o are prioritized for real-robot execution.

To obtain calibrated success probabilities, we train a Gaussian Process (GP) classifier over SE(3) on the LLM-provided labels $\{(\mathbf{t}_i,\mathbf{q}i),y_i\}$. At test time, the GP outputs calibrated success probabilities. We use the predicted probabilities to rank all candidate strategies and select the one with the highest likelihood of success for real-robot execution. 

%% file: Content/results.tex
\section{results}

\subsection{Experiments Setup}
Our experiments contain two parts. First, we tested the digital twin alignment accuracy.
The second part was nine different real-world manipulation tasks. The selected tasks are designed to validate the system’s ability to understand both semantic instructions and spatial relations, and also the ability to complete various open-world manipulation tasks without seeing the objects or demonstrations before.

\begin{table*}[h]
\centering
\caption{Comparison of mesh alignment performance between our two-stage alignment pipeline and direct alignment.}
\small
\begin{tabular}{p{2.8cm}|p{2cm}|p{1.8cm}|p{1.5cm}|p{1.5cm}|p{1.5cm}}
\hline
\multirow{2}{*}{\textbf{objects}}  & \multirow{2}{*}{\textbf{valid sample}} & \multicolumn{2}{c|}{\textbf{our method}} & \multicolumn{2}{c}{\textbf{direct alignment}} \\
\cline{3-6}
& & success rate ↑ & RMSE ↓ & success rate & RMSE \\
\hline
\multicolumn{6}{l}{\textbf{without grasp in the gripper}} \\
\hline
mug & 11 & \textbf{90.91\%} & 0.00457 & 27.27\% & \textbf{0.00390} \\
box & 5  & \textbf{100.00\%} & \textbf{0.00564} & \textbf{100.00\%} & 0.00684 \\
foam box container & 5  & \textbf{40.00\%} & \textbf{0.00665} & 20.00\% & 0.00745 \\
bottle & 8  & \textbf{100.00\%} & 0.00432 & 62.50\% & \textbf{0.00367} \\
laptop & 11  & \textbf{90.91\%} & \textbf{0.00408} & 45.45\% & 0.00453 \\
\hline
\multicolumn{6}{l}{\textbf{grasped in the gripper}} \\
\hline
mug & 9 & \textbf{88.89\%} & \textbf{0.00443} & 11.11\% & 0.00450 \\
banana & 8 & \textbf{100.00\%} & 0.00505 & \textbf{100.00\%} & \textbf{0.00460} \\
fish & 4 & \textbf{100.00\%} & 0.00400 & \textbf{100.00\%} & \textbf{0.00317} \\
cup & 3 & \textbf{66.67\%} & \textbf{0.00412} & 33.33\% & 0.00474 \\
\hline
\end{tabular}
\label{tab:mesh_alignment}
\end{table*}

\begin{figure}[h]
    \centering
    \includegraphics[width=0.9\linewidth]{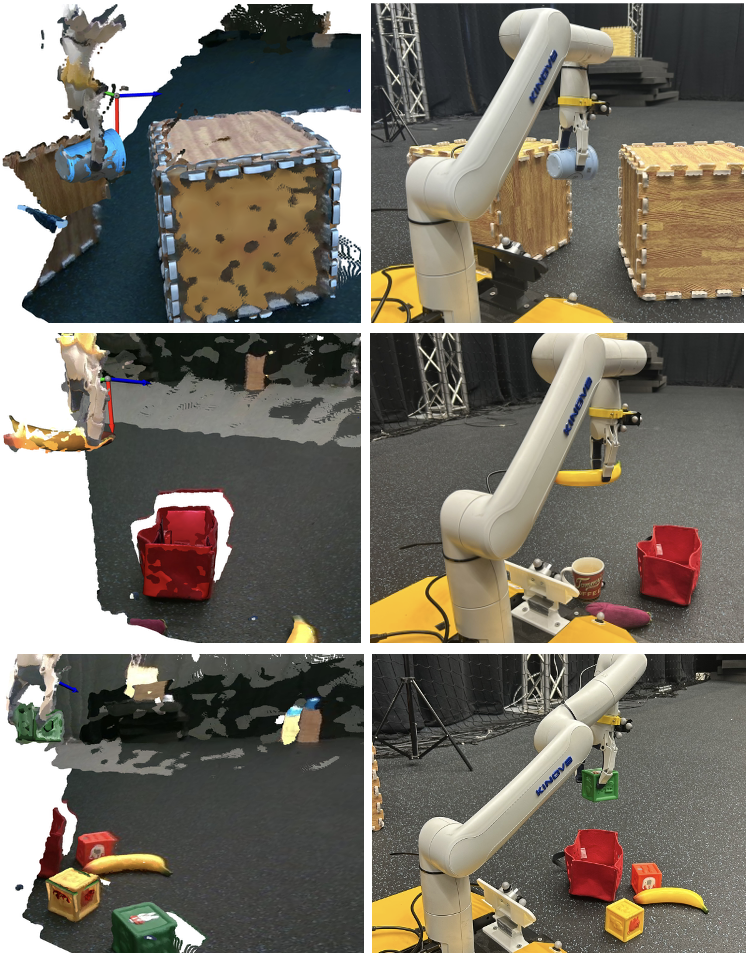}
    \caption{Qualitative results of the proposed two-stage mesh alignment method.
    The left image shows the aligned meshes overlaid with the RGB-D observation.
    The right image illustrates the corresponding real-world setup.}
    \label{fig:mesh_alignment}
\end{figure}

We used the 7-DoF Kinova Gen3 Lite robotic arm.
Two Intel RealSense D405 stereo cameras
were mounted on the gripper and the front bottom on the mobile base, 
respectively. 
The bottom camera 
has a 25 degree elevation angle to better observe the grasped object. For computational resources, we have an NVIDIA RTX 2070 GPU with 6GB RAM on a laptop for Isaac Sim simulation, and an NVIDIA RTX 3090 GPU workstation with 24GB RAM for computer vision models like Grounded-SAM and Hunyuan 3D 2.0.

\subsection{Digital Twin Alignment Accuracy}\label{sec:alignment_exp}
In this experiment, we compare our proposed two-stage mesh alignment pipeline with a direct alignment baseline, which applies RANSAC and ICP directly to the generated mesh. Ideally, the pose error of each object would serve as a more informative metric for evaluating alignment performance; however, ground-truth object poses are unavailable in the real-world setup, and coordinate frames are not consistently fixed for each generated mesh. Therefore, we evaluate alignment performance using two complementary metrics: alignment success rate, which measures the correctness of the estimated object pose, and root mean square error (RMSE), which assesses the geometric precision of the alignment.

We conduct experiments across multiple object categories, both being placed freely and under grasping scenarios. Since ground-truth meshes are also unavailable, we use Hunyuan 3D to generate meshes for all samples. Both the proposed and baseline alignment pipelines use the same generated meshes for a fair comparison. The qualitative results are shown in Fig.~\ref{fig:mesh_alignment}, and quantitative results are summarized in Table~\ref{tab:mesh_alignment}.

The results demonstrate that our proposed two-stage mesh alignment method substantially improves the alignment success rate compared with direct alignment using RANSAC and ICP. The direct alignment baseline relies solely on minimizing point-to-point distances without a reliable initialization, which often leads to convergence to local minima or incorrect correspondences.
In contrast, our method leverages an initial coarse alignment stage guided by appearance similarity, effectively narrowing the search space for fine alignment and improving robustness to noise and scale differences. Overall, our method achieves a much higher alignment success rate while maintaining comparable geometric accuracy, validating the effectiveness of the two-stage pipeline for robust and generalizable mesh-to-scene alignment.

\begin{figure*}
    \centering
    \includegraphics[width=0.9\linewidth]{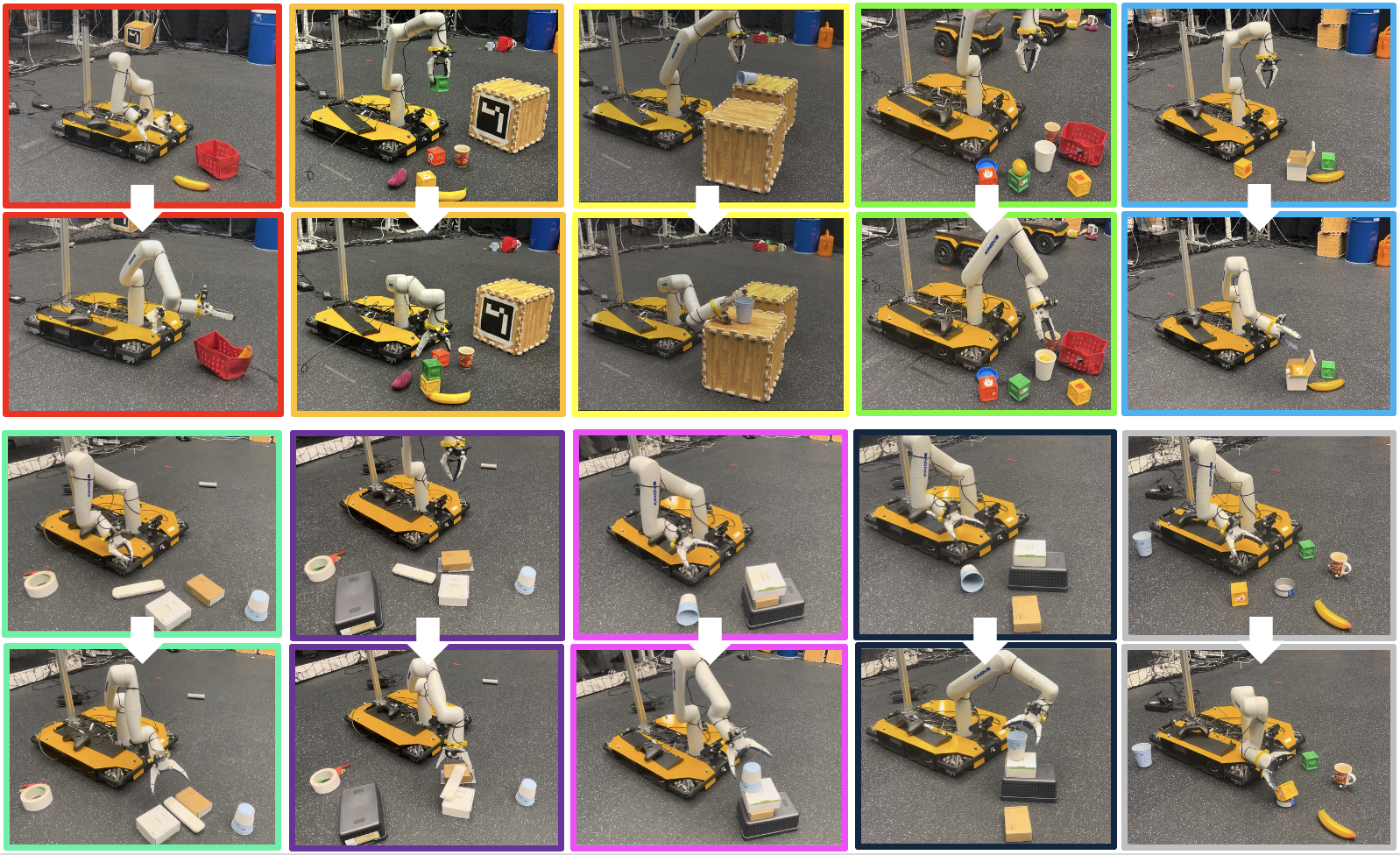}
    \caption{Representative tasks used for experimental validation. Frames with the same color represent the same task, where the top one denotes the initial scene and the bottom one denotes the final scene.}
    \label{fig:tasks}
\end{figure*}

\subsection{Real Robot Task Performance}
We performed 9 real robot experiments for a total 96 attempts on our system, as shown in Fig.~\ref{fig:tasks}. The selected tasks are designed to validate the system’s ability to understand both semantic instructions and spatial relations, and also the ability to complete various open-world tasks without seeing the objects before or fine-tuning on these tasks. Relying on these properties, our system shows the potential to tackle the open-world object manipulation problem in a novel way.

The success rates for the selected tasks are reported in Table~\ref{tab:task_success}. We observe that six of the nine tasks reach a success rate of at least 75\%. The other three tasks perform less reliably, and we discuss possible reasons in the later failure analysis.

\begin{table}[h]
\centering
\caption{Task success rate of different manipulation scenarios.}
\begin{tabular}{p{6.3cm}p{1.8cm}}
\hline
\textbf{Task} & \textbf{Success Rate} \\
\hline
Put the banana into the basket & 83.3\% (10/12) \\
Put the lemon into the white cup & 75.0\% (9/12) \\
Put the yellow cube into the white box & 91.7\% (11/12) \\
Put the yellow cube into the blue can & 75.0\% (9/12) \\
Stack the green cube onto the yellow cube & 50.0\% (6/12) \\
Put the long cutlery box on the two boxes, like a bridge & 75.0\% (9/12) \\
Put the long cutlery box into the gap between two boxes & 83.3\% (10/12) \\
Put the blue cup upside on the wooden box & 33.3\% (2/6) \\
Put the blue cup upside down on the wooden box & 33.3\% (2/6) \\
\hline
\end{tabular}
\label{tab:task_success}
\end{table}

\begin{figure}[h]
    \centering
    \includegraphics[width=1.0\linewidth]{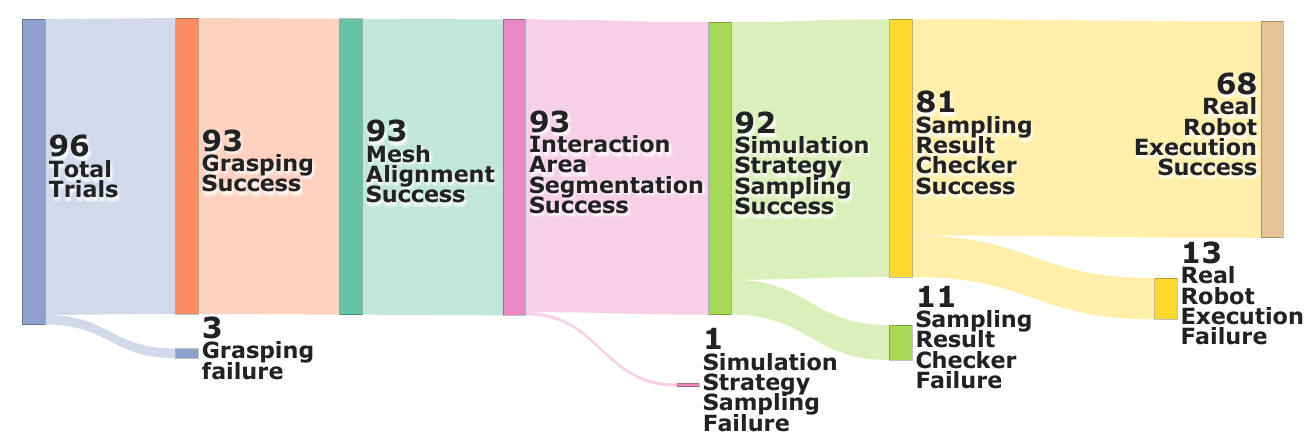}
    \caption{Sankey diagram illustrating the stage-wise success and failure flow across all 96 real-world trials. Each block represents one functional module in the proposed framework.}
    \label{fig:sankey}
\end{figure}

To provide a comprehensive evaluation of the entire system, we record the success/failure results of all major components, including object grasping, mesh alignment, interaction area segmentation, strategy sampling in simulation, sampling result checking, and real robot execution. The results are visualized as a Sankey diagram as shown in Fig~\ref{fig:sankey}.

\begin{figure}[h]
    \centering
    \includegraphics[width=\linewidth]{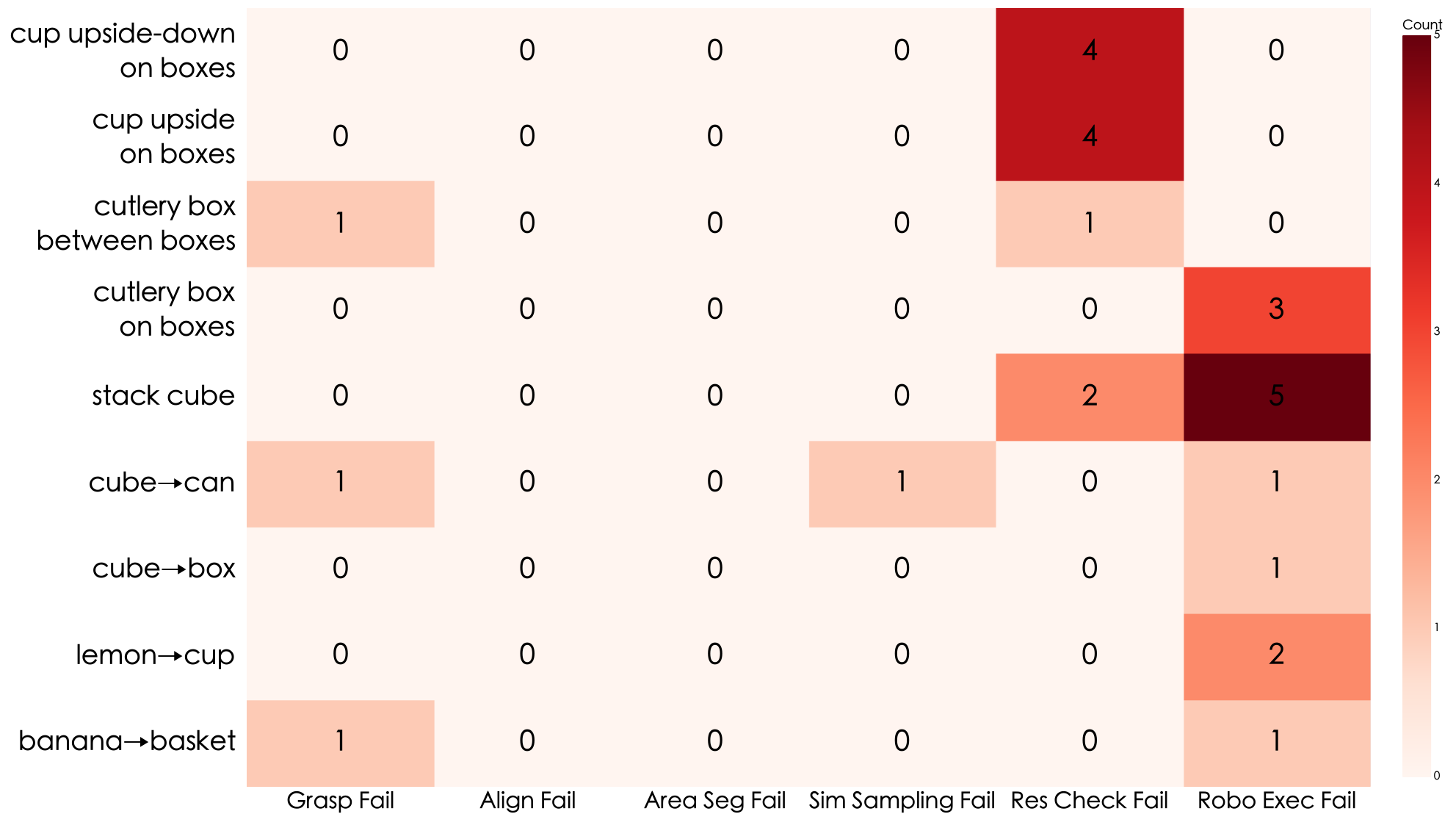}
    \caption{Failure type distribution across all 9 tasks. Each cell represents the number of occurrences of a specific failure type for a given task.}
    \label{fig:failure-type}
\end{figure}

Most of the failures came up from the result checker and final execution stage. The result checker failure denotes that the visual information were not clear enough, at least for an LLM, to make correct judgments. We also conducted a task-level failure analysis as illustrated in Fig.\ref{fig:failure-type}, and it shows that most of the result checker failures happened in the "cup on/upside down on the boxes", since the visual features of a cup standing upside and upside down were ambiguous. Overall, the LLM-based checker did demonstrate strong generalization and semantic understanding across different object configurations. However, its sensitivity to visual ambiguity suggests that incorporating multimodal feedback (e.g., depth or contact information) could further improve its robustness.

The real robot execution failure means that although the action candidates were successful in the simulator, the real robot execution still failed, indicating the sim-to-real gap was nontrivial. Several factors contribute to this discrepancy. First, the physics engine of Isaac Sim, though highly realistic, cannot perfectly replicate real-world contact dynamics. In an attempt to mitigate this gap, we incorporated material property estimation to better parameterize physical attributes in Isaac Sim. However, the range of adjustable parameters remains limited for rigid body objects. In particular, hollow or deformable structures, such as plastic cubes in the 'Stack cubes' task, cannot yet be accurately modeled or detected based solely on visual appearance, leading to further inconsistencies between simulated and real interactions. Second, the robot hardware precision and control latency introduce minor pose and timing errors that can accumulate during execution, especially for tasks with tight spatial constraints. Third, sensor noise and calibration errors (e.g., camera extrinsics and depth inaccuracies) can cause small misalignment between the reconstructed scene and the actual setup.

%% file: Content/discussion.tex
\section{Conclusion}
This work presented a novel framework that constructs a mesh-based explicit world model to enable action sampling and evaluation for open-world manipulation tasks. Our approach leverages VLMs and a simulator to reason about object dynamics and choose successful actions without any demonstrations.
Experimental results demonstrate that the proposed two-stage mesh alignment pipeline is able to support the accuracy and robustness of digital twin reconstruction, providing more reliable object pose estimation.
With the constructed explicit world model, the system is capable of performing dynamic reasoning and strategy evaluation across diverse manipulation tasks, and achieves successful transfer from simulation to real-world execution, proven by real robot experiments.

While the proposed framework demonstrates promising results, the simulation-based strategy sampling process remains computationally demanding, which currently prevents real-time deployment. 
Future work will focus on improving computational efficiency to facilitate efficient real-world deployment.